%% file: main.tex
\documentclass[letterpaper]{article} 
\usepackage[draft]{aaai25}  
\usepackage{times}  
\usepackage{helvet}  
\usepackage{courier}  
\usepackage[hyphens]{url}  
\usepackage{graphicx} 
\usepackage{natbib}  
\usepackage{caption} 
\usepackage{algorithm}

\usepackage{booktabs}
\usepackage{subcaption}
\usepackage{array}
\usepackage{multirow}
\usepackage{amsmath}
\usepackage{algpseudocode}
\usepackage{xcolor}

\usepackage{newfloat}
\usepackage{listings}
\DeclareCaptionStyle{ruled}{labelfont=normalfont,labelsep=colon,strut=off} 
\floatstyle{ruled}
\newfloat{listing}{tb}{lst}{}
\floatname{listing}{Listing}
\title{VLMine: Long-Tail Data Mining with Vision Language Models}
\author{
    Mao Ye\textsuperscript{\rm 1},
    Gregory P. Meyer\textsuperscript{\rm 1},
    Zaiwei Zhang\textsuperscript{\rm 1},
    Dennis Park\textsuperscript{\rm 1},
    Siva Karthik Mustikovela\textsuperscript{\rm 1},
    Yuning Chai\textsuperscript{\rm 2},
    Eric M Wolff\textsuperscript{\rm 1}
}

\affiliations{
    \textsuperscript{\rm 1}Cruise LLC.\\
    \textsuperscript{\rm 2}Meta Inc.\\

}

\usepackage{bibentry}

\begin{document}

\maketitle

\begin{abstract}
Ensuring robust performance on long-tail examples is an important problem for many real-world applications of machine learning, such as autonomous driving. This work focuses on the problem of identifying rare examples within a corpus of unlabeled data. We propose a simple and scalable data mining approach that leverages the knowledge contained within a large vision language model (VLM). Our approach utilizes a VLM to summarize the content of an image into a set of keywords, and we identify rare examples based on keyword frequency. We find that the VLM offers a distinct signal for identifying long-tail examples when compared to conventional methods based on model uncertainty. Therefore, we propose a simple and general approach for integrating signals from multiple mining algorithms. We evaluate the proposed method on two diverse tasks: 2D image classification, in which inter-class variation is the primary source of data diversity, and on 3D object detection, where intra-class variation is the main concern. Furthermore, through the detection task, we demonstrate that the knowledge extracted from 2D images is transferable to the 3D domain. Our experiments consistently show large improvements (between 10\% and 50\%) over the baseline techniques on several representative benchmarks: ImageNet-LT, Places-LT, and the Waymo Open Dataset.
\end{abstract}

%

\input{tex/introduction}

\input{tex/related}
\input{tex/method}
\input{tex/experiment}

\input{tex/conclusion}

\bibliography{main}

\clearpage
\appendix
\input{tex/appendix}

\end{document}

%% file: tex/introduction.tex
\section{Introduction}
\label{sec:intro}
In many real-world robotic applications, such as autonomous driving, the model must be able to handle a wide variety of situations. Ensuring the model's performance in long-tail scenarios remains a challenging problem.
Previous efforts on long-tail learning include improving model optimization \cite{cui2019class_loss_reweighting, cui2021parametric}, model modification \cite{menon2020long, wang2020long}, and fine-tuning from pre-trained models \cite{long2022retrieval, shi2023parameter}. While these works significantly improve the model's performance on long-tail data, they mainly focus on how to improve the model given a fixed dataset. Alternatively, this paper focuses on mining for long-tail examples that can be added a training set to improve model performance. In many real-world applications, there exists a relatively small amount of labeled data and a vast amount of unlabeled data. Take autonomous driving for example, the vehicles collect enormous quantities of raw sensor data while driving. Most of this data is uninteresting; thus, annotating it would be costly without providing much benefit. However, a tiny portion of the data consists of interest long-tail situations unavailable in our existing training set. As a result, automatic mining for this long-tail data is of the utmost importance.

{\color{black} Identifying long-tail examples from a large pool of data can be challenging. Existing approaches primarily utilize model uncertainty as a key signal \cite{gal2017deep, jiang2022improving, choi2021active}. The underlying assumption is that the model tends to be less confident in its predictions for long-tail examples. However, an example that results in a high model uncertainty might not be a long-tail example but a hard example. For instance, in the study by \cite{jiang2022improving} on LiDAR-based outdoor 3D detection, hard examples generally consist of long-range or occluded vehicles. These hard examples are common, yet the model produces highly uncertain predictions for them, which significantly reduces the effectiveness of using uncertainty as a criterion for selecting long-tail examples.}

Recently, foundation models such as large language models (LLMs) \cite{chowdhery2023palm,touvron2023llama,openai2023gpt,chiang2023vicuna} and large vision language models (VLMs) \cite{alayrac2022flamingo,dai2023instructblip,awadalla2023openflamingo,liu2024visual,zhu2024minigpt} have emerged. These models are trained with internet-scale data and have demonstrated impressive few-shot and zero-shot performance on recognition tasks \cite{radford2021learning, liu2024visual}. Leveraging the knowledge of VLM also benefits other tasks such as image clustering \cite{kwon2024image} and out-of-domain detection \cite{jiang2024negative}.

{\color{black}
We argue that VLMs offer a more comprehensive understanding of long-tail examples due to their rich semantic extraction capabilities. However, the challenge lies in leveraging VLMs to enhance the performance of task-specific models on these examples. Directly deploying VLMs is often impractical due to latency concerns and the complexity of adapting them to task with different modalities. This paper demonstrates that the knowledge embedded in VLMs can serve as an effective signal for mining long-tail examples, thereby improving the quality of training data and enhancing the performance of task-specific models. The intuition is straightforward: since VLMs have been exposed to a highly diverse set of examples, they are capable of providing a more comprehensive understanding of semantic information. By comparing the semantic information described by the VLM, we can identify examples with less frequent semantic descriptions, which serves as a stronger signal for detecting long-tail instances. To measure the rareness of an example based on these descriptions, we employ a simple keyword-based approach that summarizes the description into a set of keywords. The rareness of an example can then be approximated by the frequency of these keywords. We name our approach VLMine, and to the best of our knowledge, this is the first work to utilize a VLM for data mining.} 

VLMine does not use any information from the task-specific model. For that reason, the long-tail signal provided by our approach tends to be complementary to the signals obtained from traditional model-based mining algorithms \cite{beluch2018power, choi2021active, jiang2022improving}. To leverage both types of mining techniques, we propose an algorithm, referred to as Pareto mining, to integrate long-tail signals from any number of sources.

The contributions of this work are summarized below. 
\begin{itemize}
    \item We propose VLMine, a simple model-agnostic long-tail data mining algorithm that leverages the world knowledge from a VLM.
    \item Additionally, we propose an algorithm for integrating the long-tail signal from both model-based and model-agnostic mining techniques.
    \item We evaluate the data mining performance with end-to-end metrics on 2D image classification and 3D object detection benchmarks, which demonstrates the ability of our proposed method to identify both inter- and intra-class long-tail variations on multiple domains. We show consistent improvement over baselines methods on ImageNet-LT, Places-LT, and the Waymo Open Dataset.
\end{itemize}

%% file: tex/related.tex
\section{Related Work} \label{sec: related work}
\paragraph{Long-tail Perception}
Most existing work on long-tail visual recognition considers objects with less frequent class labels as long-tail examples and develops approaches to improve prediction accuracy in such label-imbalance situations. Applications include 2D tasks such as image classification \cite{cui2019class, kang2019decoupling, liu2019large}, segmentation \cite{hsieh2021droploss, wang2021seesaw, wu2020forest}, and object detection \cite{li2020overcoming, tan2020equalization, tan2021equalization, hyun2022long}. Typical approaches for long-tail perception include data reweighting and resampling \cite{cui2019class_loss_reweighting, cao2019learning, khan2019striking, huang2019deep, zhang2021distribution, hyun2022long, chawla2002smote, han2005borderline}, gradient balancing \cite{tang2020long, tan2021equalization_gradient_balance}, model or logits modifications \cite{wang2020long, alshammari2022long, menon2020long, ren2020balanced}, knowledge transfer or distillation \cite{wang2017learning, chu2020feature, kim2020m2m, liu2021gistnet, xiang2020learning, li2021self}, representation learning \cite{zhong2021improving, cui2021parametric, du2024probabilistic}, and fine-tuning pre-trained models \cite{ma2021simple, tian2022vl, dong2022lpt, long2022retrieval, shi2023parameter}. 

Compared with 2D vision, long-tail 3D object detection is less explored. \cite{peri2023towards} developed feature sharing and camera-LiDAR fusion mechanisms to improve 3D detection on objects with infrequent class labels. \cite{ma2023long} proposed a late-fusion approach of camera and LiDAR features for 3D detection on uncommon classes. Both \cite{peri2023towards, ma2023long} focus on inter-class long-tail examples. \cite{sick2023adaptiveshape} used shape-aware anchor distributions and heatmaps combined with camera-LiDAR fusion to improve 3D detection on objects with less frequent shapes. Those approaches aim at improving the detection on long-tail examples by modifying the model architecture rather than mining long-tail data. \cite{jiang2022improving} was the first to study intra-class long-tail 3D detection and proposed a density-based (using a flow model) and uncertainty-based approach for mining long-tail examples. Compared with \cite{jiang2022improving}, we consider a new data mining approach by leveraging the knowledge from a pre-trained VLM.

\paragraph{Active Learning}
This work is closely related to pool-based active learning, where the goal is to actively select samples from a large set of unlabeled data to be labeled. Current active learning approaches can be classified into two categories: uncertainty-based and density-based. Uncertainty-based approaches use a model's level of uncertainty as a signal to identify hard examples. The methods to measure uncertainty vary, with some representative approaches including ensemble variance \cite{beluch2018power, choi2021active}, entropy of the predicted distribution \cite{holub2008entropy, segal2022just}, and Bayesian modeling \cite{gal2017deep, harakeh2020bayesod} of the logits. The other category selects examples such that the data distribution formed by those examples closely approximates the population data distribution \cite{sinha2019variational, sener2017active, gudovskiy2020deep}. These approaches are not designed specifically for long-tail data mining. Despite some close connections between long-tail data mining and active learning, the goals are different. Active learning aims to reduce labeling costs by properly selecting a smaller amount of data, while data mining aims to discover long-tail data. To the best of our knowledge, we are the first to utilize a VLM to select data for labeling.



%% file: tex/method.tex
\section{Method}
\begin{figure*}[t]
\centering
\includegraphics[width=\linewidth]{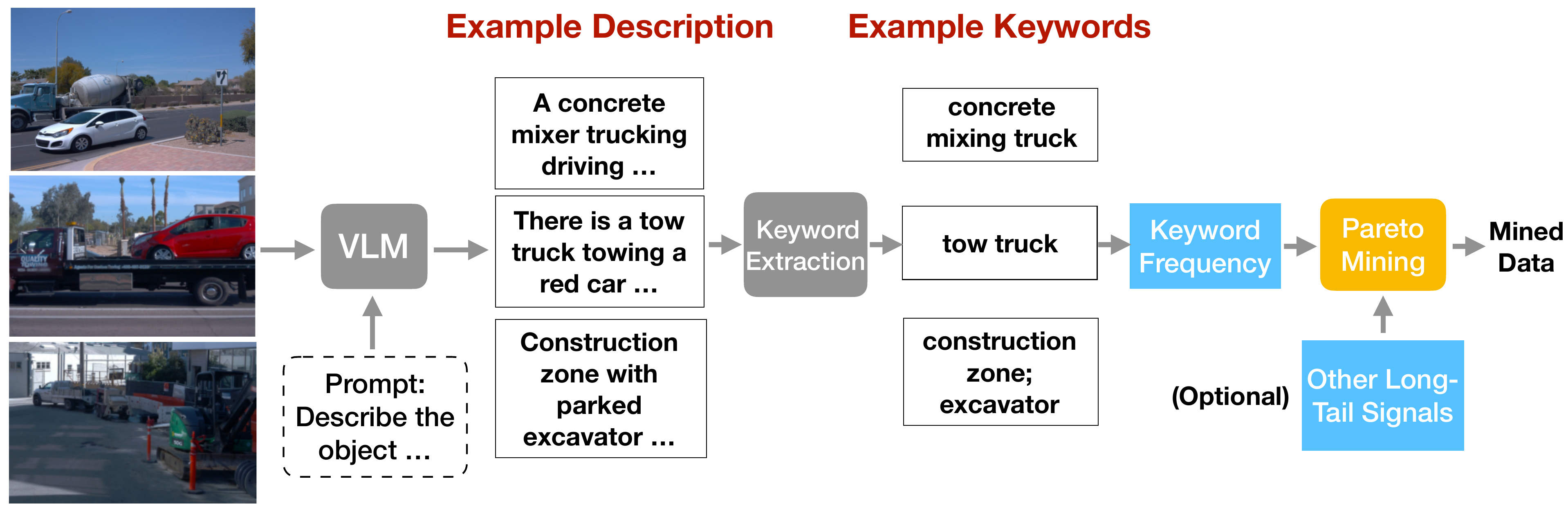}
\caption{An overview of our proposed method. First, we prompt a VLM to describe the image, then the descriptions are summarized into a set of representative keywords using a rule-based heuristic or LLM. The frequency of the keywords are used to score the novelty of the images. Afterwards, the score can be combined with other long-tail signals, and our proposed Pareto mining is used to select the long-tail data to be labeled.}
\label{fig:illustration}
\end{figure*}

Our goal is to find long-tail examples from a large pool of unlabeled data. Depending on the problem and data, the long-tail examples can be either inter-class, i.e. less common class categories, or intra-class, i.e. less common variations within a class. The main idea of our approach is to leverage the world knowledge from a VLM and LLM to extract keywords that describe the examples, and the frequency of the keywords is used to select the data to be labeled. Furthermore, our proposed Pareto mining can be used to incorperate other long-tail signal. An overview of our proposed method is illustrated in Figure \ref{fig:illustration}.

\subsection{VLMine: VLM Knowledge for Long-tail Example Mining}
VLMine is a straightforward mining procedure consisting of two steps: extraction of representative keywords and example selection via keyword frequency.

\paragraph{Extraction of representative keywords.} We summarize the examples with a set of keywords. Given an image, we first prompt the VLM to describe the image, and then we summarize the description into a set of keywords. This can be done using a rule-based approach or by querying a LLM. After extracting the keywords, we also apply some standard post-processing, such as lemmatization and stop word removal. Note that each example might have a different number of keywords. 


\paragraph{Example selection via keyword frequency.} We compute the frequency of the keywords within all of the examples. The novelty score of each example is defined as a reverse monotonic transformation of the pooled keyword frequency. Depending on the data and the task, the pooling operator can be average pooling (i.e. averaging the frequency of all the keywords) or min-pooling (i.e. choosing the frequency of the least common keyword). Concretely, suppose an example has $n$ keywords with frequency $k_1, k_2, ..., k_n$, depends on the choice of pooling operator, its novelty score $s$ is defined as 
\begin{equation} \label{eq: novelty socre}
    s := \phi(\min_{i} k_i) \quad\text{or}\quad s := \phi(\textstyle{\sum_{i}} k_i / n), 
\end{equation}
where $\phi$ is any reverse monotonic transformation, e.g. $\phi(x)=-x$. Examples with higher novelty scores are assumed to be more rare. {\color{black} The choice of pooling function is task dependent. For example, in object detection, each example can contain multiple objects with most being commonly occurring objects. Thus, a majority of the keywords will be associated to common objects. In this case, the min-pooling operator is a better choice. However, for image classification, we expect most of the keywords to be associated with the class and thus average pooling becomes a more reasonable choice.}

We could directly ask the VLM to identify examples that are less common. However, such an approach selects examples based on the VLM's knowledge, which introduces bias. The examples that the VLM believes to be uncommon might not actually be rare in the task database.

\subsection{Pareto Mining: Combining Multiple Data Mining Signals}
The novelty score determined by VLMine is model-agnostic. It purely uses the knowledge from a VLM and possibly a LLM without utilizing any information from the task-specific model we want to improve, e.g. an image classifier or object detector. We find that the signal provided by VLMine is often complementary to the signals obtained from traditional model-based approaches such as uncertainty-based mining. On the other hand, uncertainty-based mining algorithms, which all rely on the task-model's knowledge, are often correlated with each other. Intuitively, if multiple signals are highly correlated, a long-tail example that is missed by one technique is likely to be missed by another. However, if the signals are complementary, even if an example is missed by one algorithm, we have a higher chance to detect it with another. Therefore, it is beneficial to develop a mining strategy that can integrated multiple signals.

\begin{figure}
  \centering
  \includegraphics[width=\linewidth]{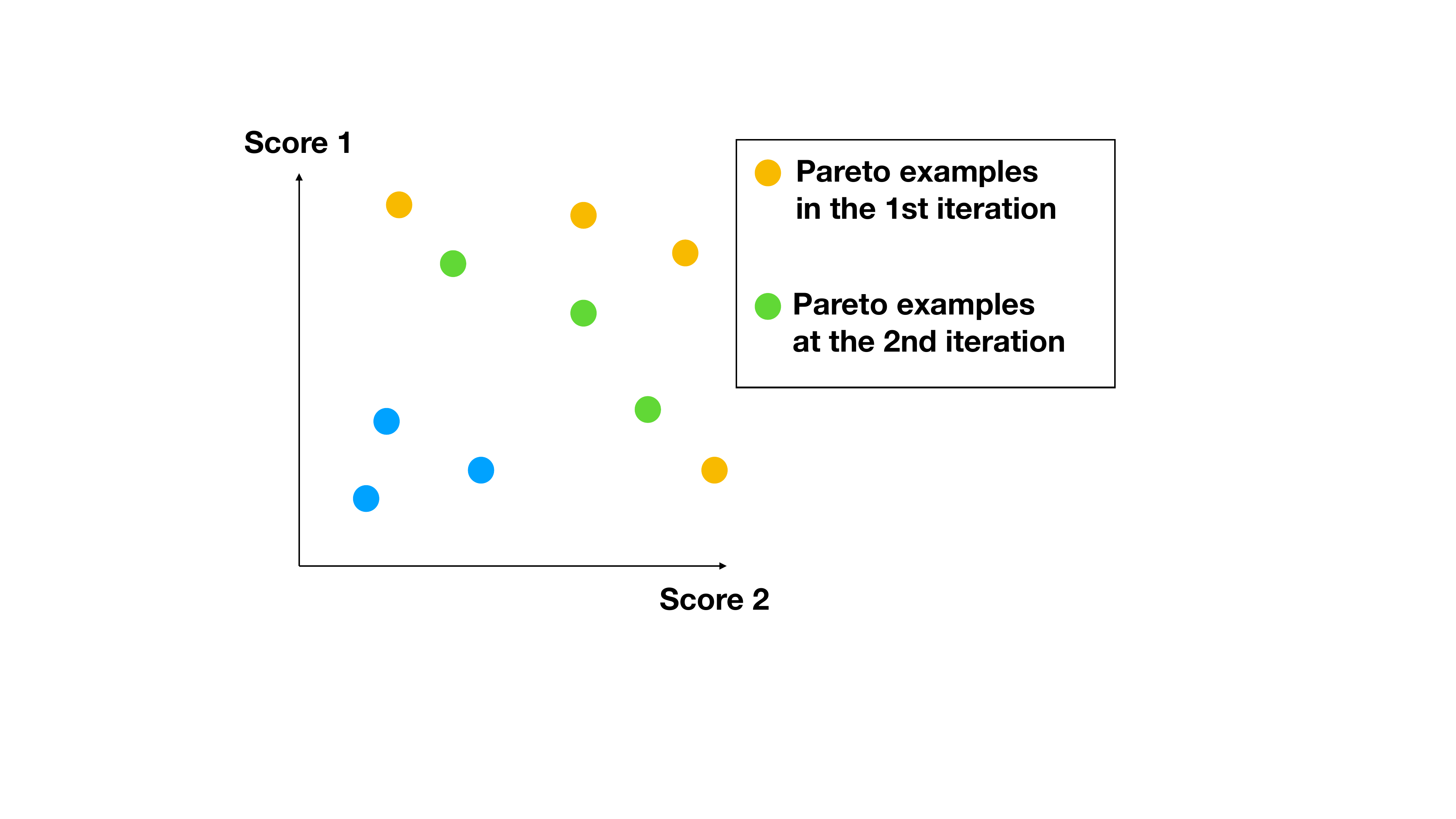}
    \caption{Illustration of the Pareto examples identified by Pareto mining.}
    \label{fig:pareto_illustration}
\end{figure}

An obvious approach is to combine the signals from multiple algorithms is to use a linear combinations of their scores. However, the final result will heavily rely on the weights of the linear combination, which may be hard to determine since the magnitude of each signal might be very different and there is not an explicit objective to optimize. Refer to Appendix \ref{appendix: linear combination} for an analysis.

We propose a simple and general approach to combine the signals from multiple algorithms. The idea is motivated by the concept of a Pareto frontier in multi-objective optimization. For each example, suppose we receive $n$ scores $\textbf{s} = [s_1, s_2, ..., s_n]$ from different algorithms that represent the novelty of the example. Without loss of generality, we assume that a signal with a higher value means a higher chance that the example is part of the long-tail. We say that an example with scores $\textbf{s}$ is dominated by an example with scores $\textbf{s}'$ if $\forall i \in {1,...,n}$, $s_i \le s'_i$ and there exists $j \in {1,...,n}$ such that $s_j < s'_j$.

We say an example with scores $\textbf{s}$ is a Pareto example if it is not dominated by any other examples. Figure \ref{fig:pareto_illustration} illustrates the case when two novelty scores are used. We add samples from the Pareto examples into the mined dataset. Once all of the Pareto examples are added, we remove them from the frontier and determine the new Pareto examples until we obtain the desired amount of mined examples. In Figure \ref{fig:pareto_illustration}, the examples colored yellow are the Pareto examples at the start of mining, and the examples colored green and blue are dominated. After the yellow Pareto examples are removed, the examples colored green become the new set of Pareto examples. See Algorithm \ref{algo: pareto mine} for the Pareto mining pseudo-code.

%% file: tex/experiment.tex
\section{Experiments} \label{sec: experiment}


Our experiments are organized as follows. In Section \ref{sec: 2d_img_experiment} and \ref{sec: 3d_det_experiment}, we evaluate VLMine and Pareto mining using a pool-based active learning setup. Our goal is to demonstrate that the proposed method is effective in identifying both inter- and intra-class long-tail examples. Details on the prompt engineering and implementations are also discussed in these sections. 
Additional visualizations, studies, and remarks are presented in the appendix.
In Appendix \ref{appendix: analyze mined example}, we provide visualizations to illustrate that VLMine successfully mines interesting long-tail examples.  In Appendix \ref{appendix: analyze pareto example}, we visualize examples identified by combining model-agnostic and model-based signals through Pareto mining. In Appendix \ref{appendix: ablation}, we conduct an ablation study to understand how the design of the prompt influences the final performance, and to compare Pareto mining against other methods for combining novelty scores. In Appendix \ref{appendix: one-step} and \ref{appendix: hallucination}, we discuss limitations related to the VLM and their impact on VLMine.




\begin{algorithm}[t]
\begin{algorithmic}
\State{\textbf{Input:}} Set of examples $\mathcal{I}$, number of examples to mine $N$
\State{\textbf{Output:}} Set of mined examples $\mathcal{M}$
\State{$\mathcal{M}\gets\emptyset$}
\While{$|\mathcal{M}| < N$}
\State{Identify the subset of Pareto examples $\mathcal{P}\subset\mathcal{I}$}
\If{$|\mathcal{P}| \le N - |\mathcal{M}|$}
    \State $\mathcal{M}\gets\mathcal{M} \cup \mathcal{P}$
\Else
    \State{Create $\hat{\mathcal{P}}$ by sampling $N - |\mathcal{M}|$ examples from $\mathcal{P}$}
    \State{$\mathcal{M}\gets\mathcal{M} \cup \hat{\mathcal{P}}$}
\EndIf
\EndWhile
\end{algorithmic}
\caption{Pseudo-code for Pareto Mining}\label{algo: pareto mine}
\end{algorithm}

\subsection{Long-tail 2D Image Classification} \label{sec: 2d_img_experiment}
\paragraph{Setup.}
We consider two long-tail datasets for the image classification task: ImageNet-LT and Places-LT. The datasets were created by sampling a subset of the original dataset following a Pareto distribution \cite{liu2019large}. ImageNet-LT is an object-centric dataset with a training split consists of 115.8K images from 1000 categories and  cardinality ranging from 5 to 1280. Places-LT is a scene-centric dataset with a training set that consists of 5 to 4980 images for each of its 365 classes for a total of 62.5K images. For both datasets, the validation and test splits are balanced and contain 20 and 100 examples per class, respectively.

We conduct pool-based active learning experiments following \cite{settles2009active, jiang2022improving}. For each class, we sub-sample 20\% of the training examples, keeping the distribution of classes the same. This subset forms the ``labeled'' pool, and the rest of the training set and the entire validation set are considered the ``unlabeled'' pool from which we will mine data. The test set remains unchanged and is used for evaluation.

We use predictive entropy \cite{shannon1948mathematical} and variation ratios \cite{freeman1965elementary} as the baselines, which have been shown to be strong acquisition functions based on the analysis in \cite{gal2017deep}. Predictive entropy, 
\begin{equation}
    -\sum_{c}p(y=c | x) \log p(y=c | x),
\end{equation}
calculates the entropy of the predicted class distribution, and higher entropy values imply a higher novelty for the example.
Likewise, the variation ratio,
\begin{equation}
    1 - \max_{c} p(y=c | x),
\end{equation}
measures the lack of confidence in a prediction, where a larger ratio is assumed to indicate a higher chance of an example being part of the long-tail.

\paragraph{Prompt and Implementation Details.}
For these experiments, we use LLaVAv1.5-7B \cite{liu2024visual} as the VLM and a heuristic to generate keywords. For ImageNet-LT, we prompt the VLM with the following: 
{\ttfamily{What are the possible classes for this image? Give three possible answers.}}
For Places-LT, the following prompt is used: 
{\ttfamily{What are the possible scene categories for this image? Give three possible answers.}}

We prompt the VLM to provide three possible answers for each example, since we find that this results in the VLM supplying more comprehensive descriptions. For an empirical analysis of prompt, please refer to Appendix~\ref{appendix: ablation}. For these experiments, we simply treat each word in the description as a keyword. Afterwards, we apply standard language post-processing, i.e. stop word removal and lemmatization, on the keywords. The novelty score of each image is computed using the average frequency of the keywords.

For ImageNet-LT, we use a ResNet-50 \cite{he2016deep} backbone and the optimization protocol from \cite{du2024probabilistic}. \cite{du2024probabilistic} uses a contrastive loss as an auxiliary loss and achieves the state-of-the-art in long-tail learning. Furthermore, we found the logit-adjustment loss \cite{menon2020long} to be more stable than the softmax loss when training on the sub-sampled dataset. The model is trained for 90 epochs with a batch size of 256, a cosine-decayed learning rate of 0.1, and SGD optimizer with 0.9 momentum. For Places-LT, we follow the protocol from \cite{liu2019large} and use an ImageNet pre-trained ResNet-152 \cite{he2016deep} as the backbone. We fine-tune the model for 30 epochs with a batch size of 256, a cosine-decayed learning rate of 0.02, a SGD optimizer with 0.9 momentum, and standard data augmentations, i.e. randomized image crops and flips. We also apply the logit-adjustment method \cite{menon2020long} to focal loss to improve learning.

\begin{figure*}[t]
\centering
\includegraphics[width=0.49\linewidth]{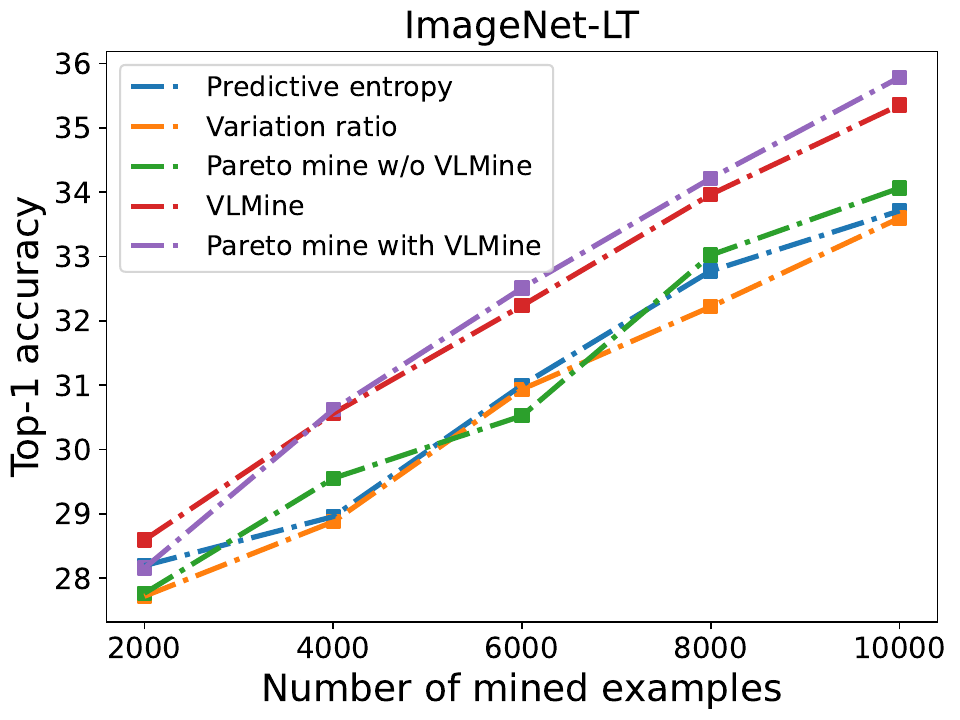}
\includegraphics[width=0.49\linewidth]{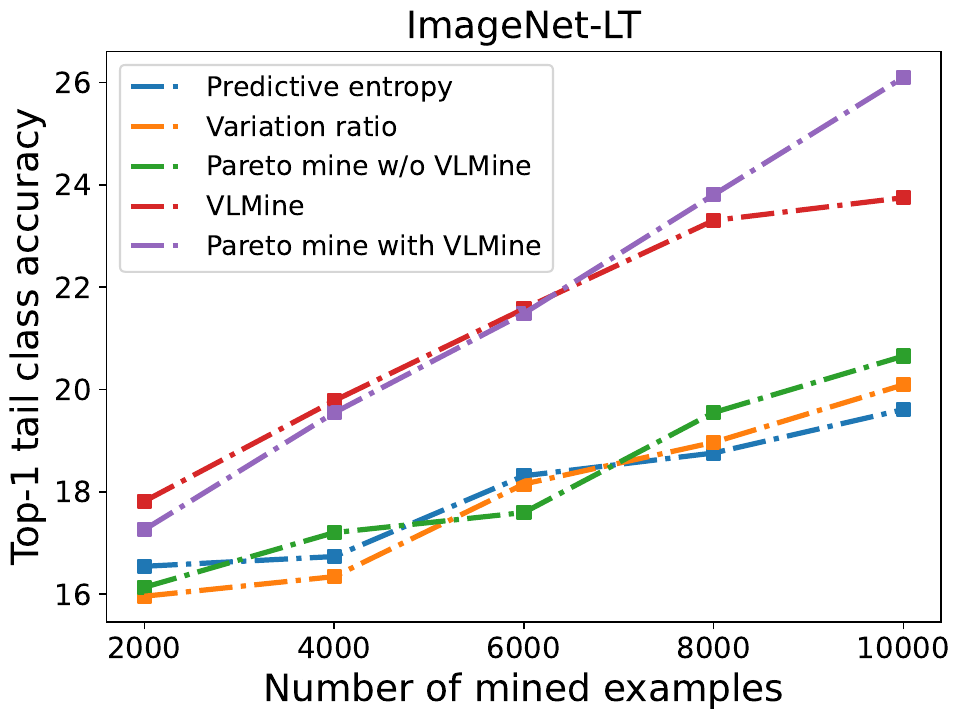}
\includegraphics[width=0.49\linewidth]{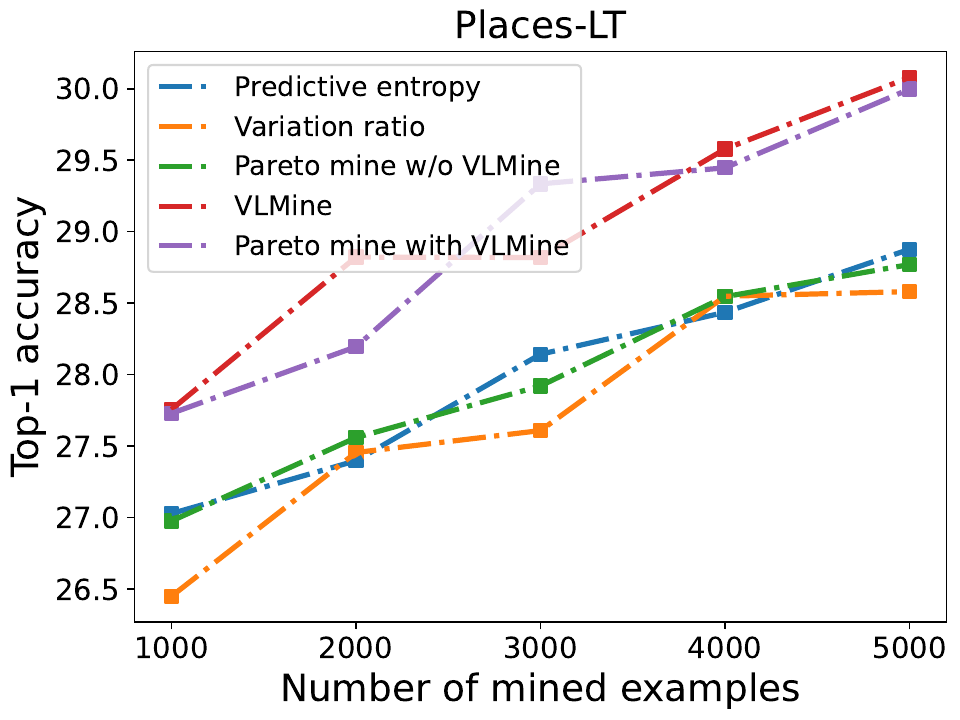}
\includegraphics[width=0.49\linewidth]{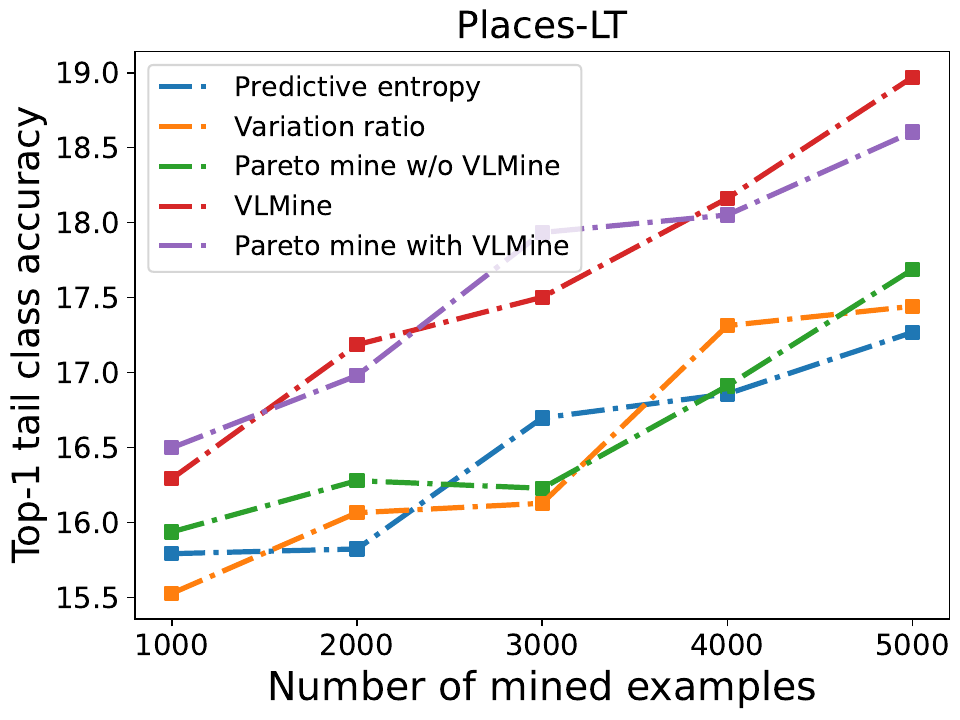}
\caption{Data mining experiments on ImageNet-LT and Places-LT.}
\label{fig:data_mine_2d_img_class}
\end{figure*}

\paragraph{Results.}
Figure \ref{fig:data_mine_2d_img_class} shows the classification accuracy on the test split for ImageNet-LT and Places-LT when different numbers of examples are mined. The number of examples mined from of the ``unlabeled'' pool varies from 10\% to 50\%. The left-hand side of Figure \ref{fig:data_mine_2d_img_class} plots the top-1 accuracy for all classes while the right-hand side plots the accuracy for only the tail classes. The model trained with data obtained from VLMine consistently outperforms the baselines. {\color{black} Note that classifying the tail classes is quite challenging as the proportion of those classes is small compared with head classes.} 

\begin{figure*}[t]
\centering
\includegraphics[width=0.45\linewidth]{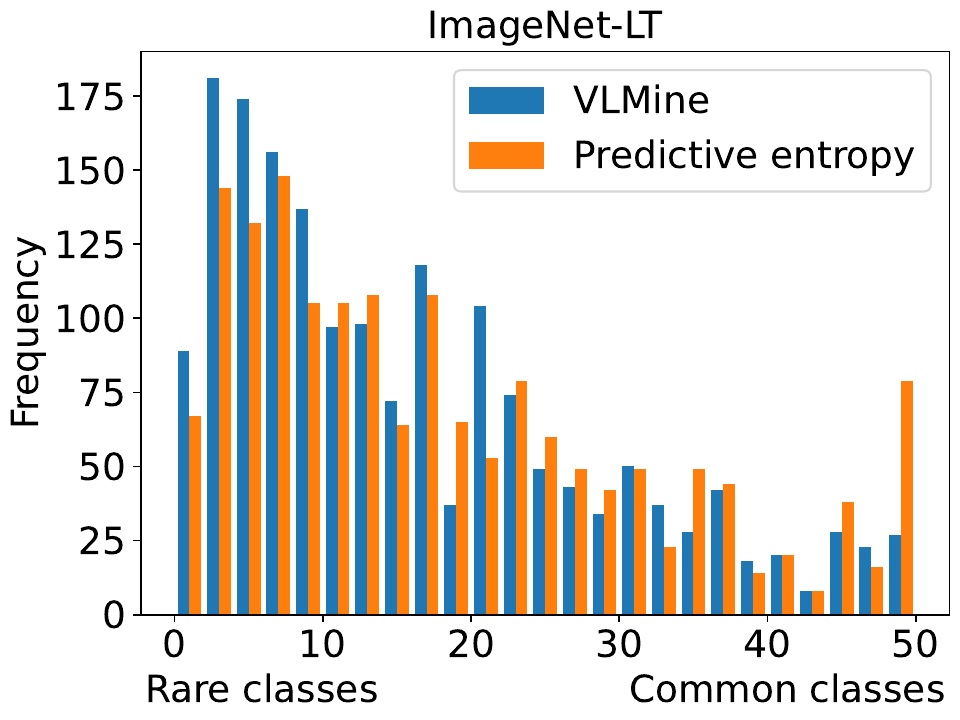}
\includegraphics[width=0.45\linewidth]{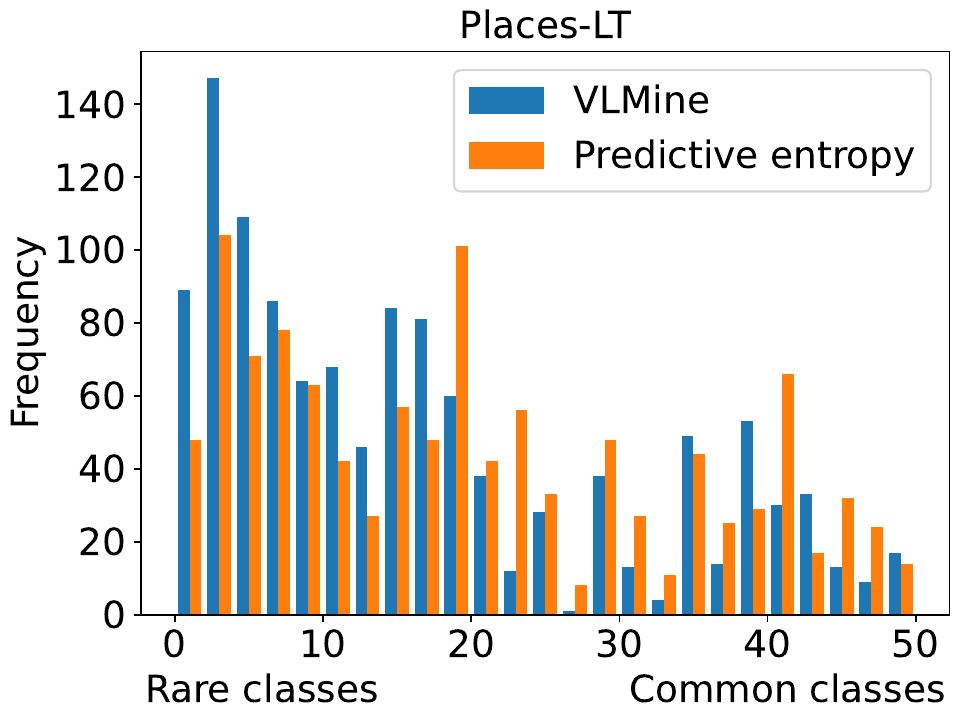}
\caption{Distribution of the mined data sorted by rareness (the bars further to the left represent rare classes while the bars on the right correspond to more common classes). For readability, we only show classes that have less than 50 images in the original ``labeled'' pool. The rareness of each class is quantified by the number of images for the class in the original ``labeled'' pool. We plot the frequency of mined examples for different rarenesses.}
\label{fig:data_mine_2d_img_class_dist}
\end{figure*}

To provide more insight, we evaluate the mining algorithms by inspecting the long-tailness of the mined examples. We define the rareness of each mined example by the number of images corresponding to its class in the original ``labeled'' pool. Thus, a technique is better at identifying long-tail examples if the mining frequency of rarer examples is higher. Figure \ref{fig:data_mine_2d_img_class_dist} shows the distribution of the 2000 examples mined by VLMine and the predictive entropy uncertainty-based approach. As we can see, VLMine mines significantly more of the rarist classes than the uncertainty-based method. Demonstrating that a VLM is more capability at identifying long-tail examples in this case.

\begin{figure}
  \centering
  \includegraphics[width=0.8\linewidth]{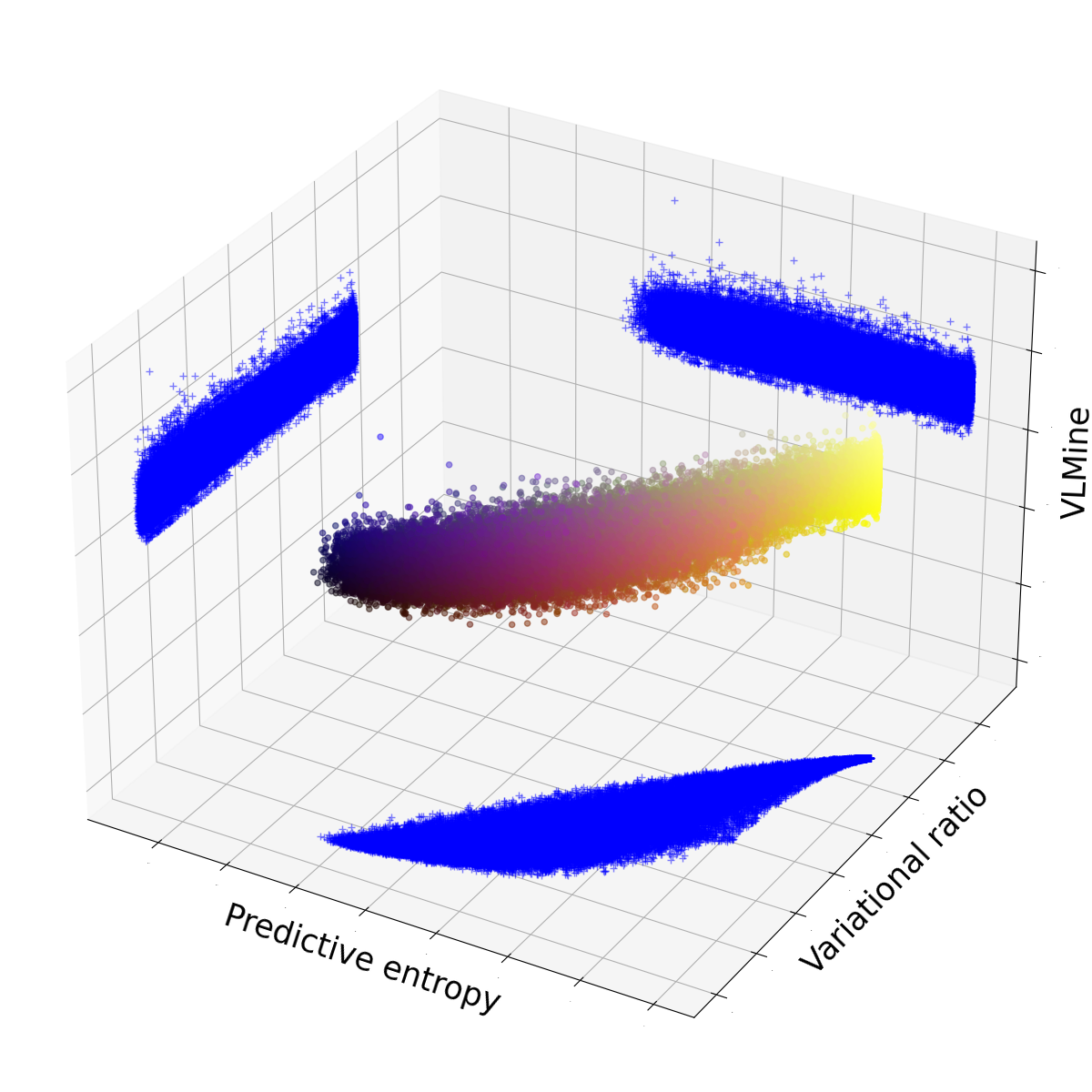}
  \caption{Correlation of the novelty scores from different algorithms on ImageNet-LT. We plot the scores from three different algorithms for each example in the unlabeled pool and project the scores to show the correlation between each pair of algorithms. As we can see, the scores between predictive entropy and variational ratio are highly correlated, while VLMine provides orthogonal signals.} \label{fig:pareto_scatter_imagenet}
\caption{Visualizations of Pareto mining.}
\end{figure}

Furthermore, we evaluate our proposed Pareto mining with and without the novelty score from VLMine and the results are shown in Figure \ref{fig:data_mine_2d_img_class}. For ImageNet-LT, Pareto mining with VLMine gives a boost in performance, especially for tail classes, when more examples are mined. In comparison, without VLMine and only using the model-based algorithms, Pareto mining does not give as significant of an improvement.
In Figure \ref{fig:pareto_scatter_imagenet}, we plot the signals from each algorithm for each example in ImageNet-LT. We obverse that the signals from the two uncertainty-based approaches are highly correlated while the signal from VLMine is orthogonal to the model-based approaches. This illustrates why combining model-agnostic and model-based signals improves data mining. 

\subsection{Long-tail 3D Object Detection} \label{sec: 3d_det_experiment}
We evaluate our proposed method on the real-world task of 3D object detection for autonomous driving. For this challenging setting, VLMine needs to identify long-tail objects within LiDAR point clouds using the corresponding camera images.

\paragraph{Setup.}
For our experiments, we use the large-scale Waymo Open Dataset \cite{sun2020scalability}. The dataset consists of 1150 sequences with each sequence containing 20 seconds of driving data captured at 10 Hz. Each frame consists of a LiDAR point cloud and 5 camera images. We follow the active learning setup from \cite{jiang2022improving}, which achieves the state-of-the-art for this task. We randomly split the sequences within the training set into a 10\% ``labeled'' pool and a 90\% ``unlabeled'' pool. The large ``unlabeled'' pool can be viewed as the stream of data coming from the autonomous fleet, from which we want to mine long-tail examples. For these experiments, the goal is to identify a small portion of the ``unlabeled'' pool (only 3\% following \cite{jiang2022improving}) that adds the most value when added to the training set.

Existing self-driving benchmarks lack fine-grained inter- or intra-class labels. For that reason, \cite{jiang2022improving} uses the shape of an object as a proxy for intra-class long-tailness. Their assumption is that a vehicle with a larger size and a pedestrian with a smaller height are part of the long-tail. Hence, they evaluate the mining techniques by decomposing the Average Precision (AP) metrics for the vehicle and pedestrian classes. Specifically, they report detection performance on larger vehicles (longer than 8 and 10 meters) and smaller pedestrians (shorter than 1.4 and 1.2 meters). Although these length/height-based decompositions are far from ideal for evaluating long-tail performance, according to the analysis in \cite{jiang2022improving}, these road users are uncommon in the Waymo Open Dataset. Our baseline for comparisons is D-REM \cite{jiang2022improving}, which is an uncertainty-based approach using an ensemble of models to infer the novelty score.

\begin{table*}[t]
\centering{}%

\begin{tabular}{c|ccc|ccc}
\multirow{2}{*}{Method} & \multicolumn{3}{c|}{Vehicle (AP/APH)} & \multicolumn{3}{c}{Pedestrian (AP/APH)}\tabularnewline
 & All & $\ge$8m & $\ge$10m & All & $\le$1.4m & $\le$1.2m\tabularnewline
\hline 
D-REM & 62.5/61.9 & 27.1/25.9 & 16.7/14.8 & \pmb{66.4/50.8} & 15.2/9.8 & 10.4/5.9\tabularnewline
VLMine & \pmb{63.0/62.3} & 38.0/37.2 & \pmb{25.2}/22.0 & 65.2/49.6 & 18.7/13.4 & 9.6/5.5\tabularnewline
Pareto Mining (VLMine + D-REM) & 62.9/62.1 & \pmb{39.3/37.6} & 25.1/\pmb{24.3} & 65.4/50.3 & \pmb{21.6/15.9} & \pmb{11.6/7.2} \tabularnewline
\end{tabular}
\caption{Data mining experiment on the Waymo Open Dataset.}
\label{tbl:active_learning}
\end{table*}

\paragraph{Prompt and Implementation Details.}
For these experiments, we again use LLaVAv1.5-7B \cite{liu2024visual} as the VLM, but generate keywords using a LLM, GPT3.5-turbo \cite{openai2023gpt}.
We prompt the VLM to describe the camera images with the following: 
{\ttfamily{Describe the uncommon or abnormal vehicles, pedestrians, and cyclists related to traffic in this image}}. Afterwards, we prompt the LLM to summarize the description into a set of keywords using: {\ttfamily{Please return keywords for each image description in this format: keyword1, keyword2, and etc}}. 
The main reason we use a LLM for keyword extraction in this case is that the situations are more complicated, and we need to ensure the keywords well represent the intra-class variations. For example, in the following description, ``There is a construction truck on the street,'' we want the algorithm to output keywords like ``construction truck'' rather than ``construction'' and ``truck''. For that reason, the heuristic approach from the previous experiment is insufficient.


We use a two-step approach to generate keywords by first asking the VLM to provide a description and then extracting keywords based on that description. Alternatively, we can directly ask the VLM to describe the example using only keywords. However, we identify some issues with that approach and refer readers to Appendix \ref{appendix: one-step} for more details.

The evaluation only considers long-tail vehicles and pedestrians, but the description from the VLM and the keywords from the LLM might contain information irrelevant to those types of objects. To address this, we filter the keywords by asking the LLM what type of road user a keyword is related to using the prompt: {\ttfamily{For each of the following words separated by ;, identify whether they are related to a description of vehicle, pedestrian, or cyclist. Example answers are: child - pedestrian; bus - vehicle; dish - not related}}.

Since each frame contains several camera images, we simply concatenate the keywords from all the images. In Section \ref{sec: 2d_img_experiment}, the images are either object-centric or scene-centric, so all the keywords are relevant to decide the rareness of the example. However, for 3D object detection, each frame may contain many objects, and the keywords describing common objects may not be relevant to the potential long-tail objects in the example. Therefore, we use the minimum frequency to compute the novelty score instead of the average frequency. This approach allows us to focus on the presence of less common keywords to determine the long-tailness of the example.

For the experiments, our detector is based on CenterPoint \cite{yin2021center}. For D-REM, we train 5 models with different random initializations while keeping all other training configurations the same. We also evaluate the integration of the two appoarches using the proposed Pareto mining. 

\paragraph{Results.}
The results of these experiments are summarized in Table \ref{tbl:active_learning}. When considering all objects, unsurprisingly all of the methods have similar performance. Our proposed VLMine significantly outperforms D-REM on all tail objects except for pedestrians under 1.2 meters where the two approaches perform comparably. We observe a further boost in performance on tail objects when employing the proposed Pareto mining technique, which demonstrates the value of combining model-agnostic and model-based novelty scores. We refer readers to Appendix \ref{appendix: analyze mined example} for a visualization of the mined examples, and to Appendix \ref{appendix: analyze pareto example} for an illustration of Pareto examples.





%% file: tex/conclusion.tex
\section{Conclusion}
This work proposes VLMine, a simple yet effective long-tail data mining algorithm that leverages the knowledge from a vision language model. VLMine provides a strong signal to detect long-tail examples which is complementary to the signals given by traditional uncertainty-based approaches. Furthermore, we show how our proposed Pareto mining can integrate multiple mining signals to further enhance performance. 

%% file: tex/appendix.tex
\section{Appendix} \label{appendix}
\subsection{Analysis of Mined Examples} \label{appendix: analyze mined example}
\begin{figure*}
\begin{subfigure}{.32\textwidth}
  \centering
  \includegraphics[width=\linewidth]{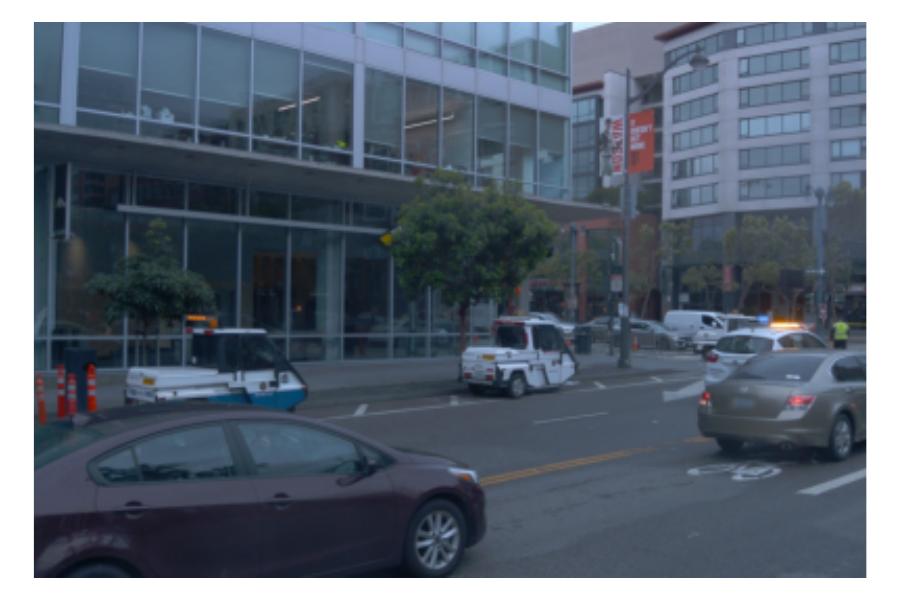}
  \caption{small trash truck}
\end{subfigure}
\begin{subfigure}{.32\textwidth}
  \centering
  \includegraphics[width=\linewidth]{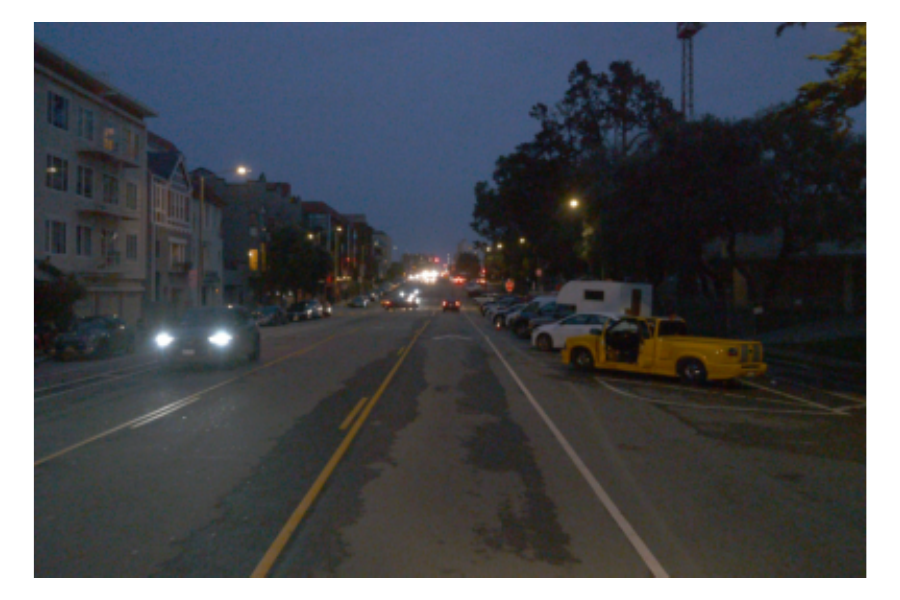}
  \caption{small school bus}
\end{subfigure}
\begin{subfigure}{.32\textwidth}
  \centering
  \includegraphics[width=\linewidth]{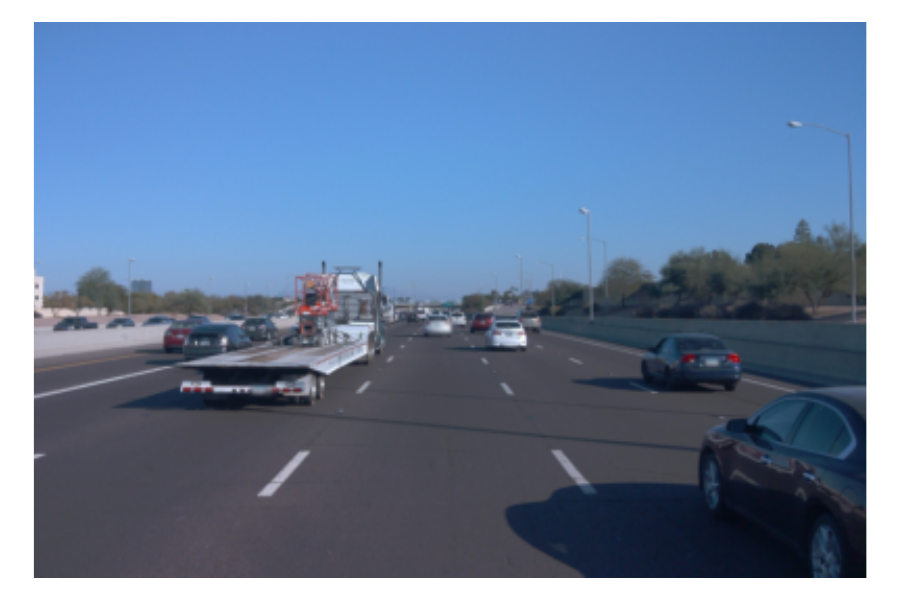}
  \caption{camouflage-patterned truck}
\end{subfigure}

\begin{subfigure}{.32\textwidth}
  \centering
  \includegraphics[width=\linewidth]{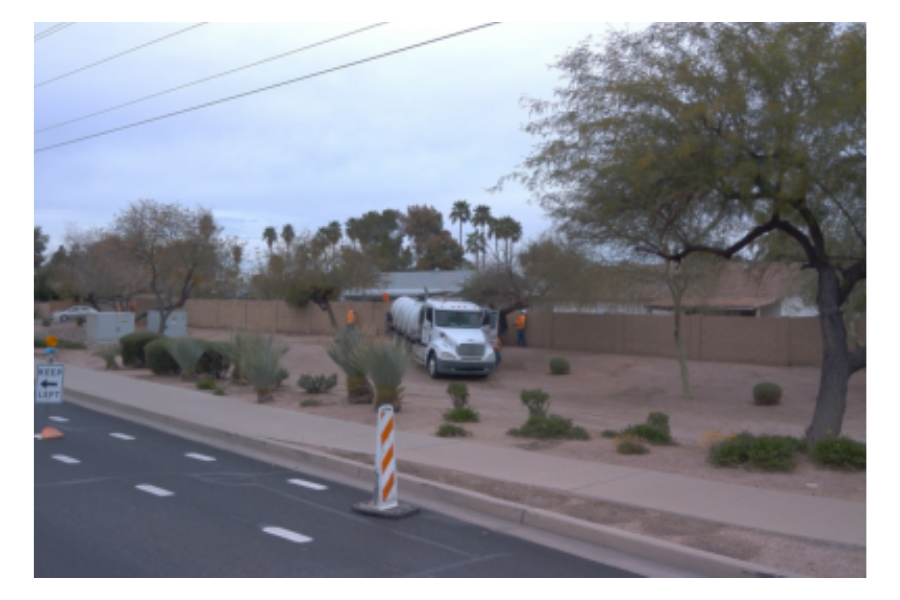}
  \caption{firefighter}
\end{subfigure}
\begin{subfigure}{.32\textwidth}
  \centering
  \includegraphics[width=\linewidth]{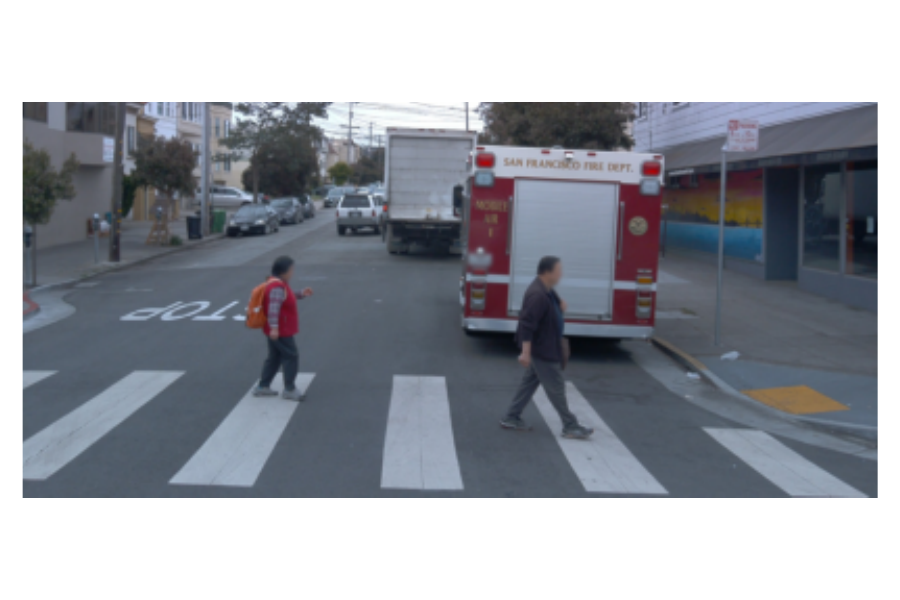}
  \caption{van parked in the middle street}
\end{subfigure}
\begin{subfigure}{.32\textwidth}
  \centering
  \includegraphics[width=\linewidth]{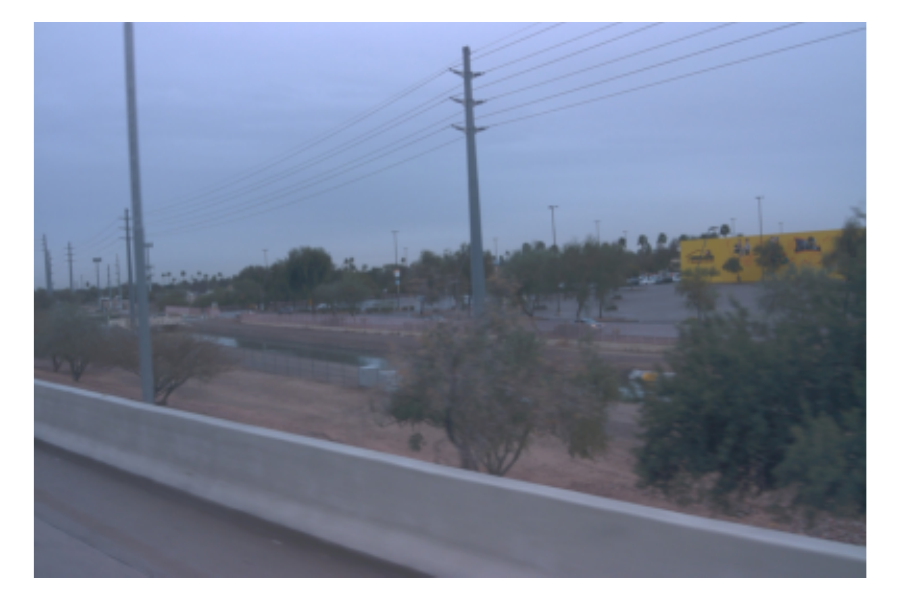}
  \caption{yellow cab truck}
\end{subfigure}
\caption{Examples of inaccurate keywords from the Waymo Open Dataset.}
\label{fig:vlm_failure}
\end{figure*}

We visualize the data mined by VLMine in Section \ref{sec: 3d_det_experiment} for the Waymo Open Dataset. Figure \ref{fig:vlm_illustration} shows the images and keywords for mined long-tail vehicles. Figure \ref{fig:vlm_failure} depicts some typical failure modes and/or less accurate keywords from VLMine. In Figure \ref{fig:vlm_failure}a-d, since the objects are hard to describe, VLMine fails to give accurate keywords. However, such failures may not degrade the quality of the mined data. The content of the images is often part of the long-tail, and although the keywords are not accurate, their frequencies are lower resulting in a higher novelty score. Figure \ref{fig:vlm_failure}e illustrates a case where VLMine detects an improperly parked van but misses the fire truck. Figure \ref{fig:vlm_failure}f shows a case where VLMine wrongly recognizes a yellow wall as a yellow truck.

\subsection{Pareto Mining Visualization} \label{appendix: analyze pareto example}
\begin{figure*}
\centering
\includegraphics[width=0.85\linewidth]{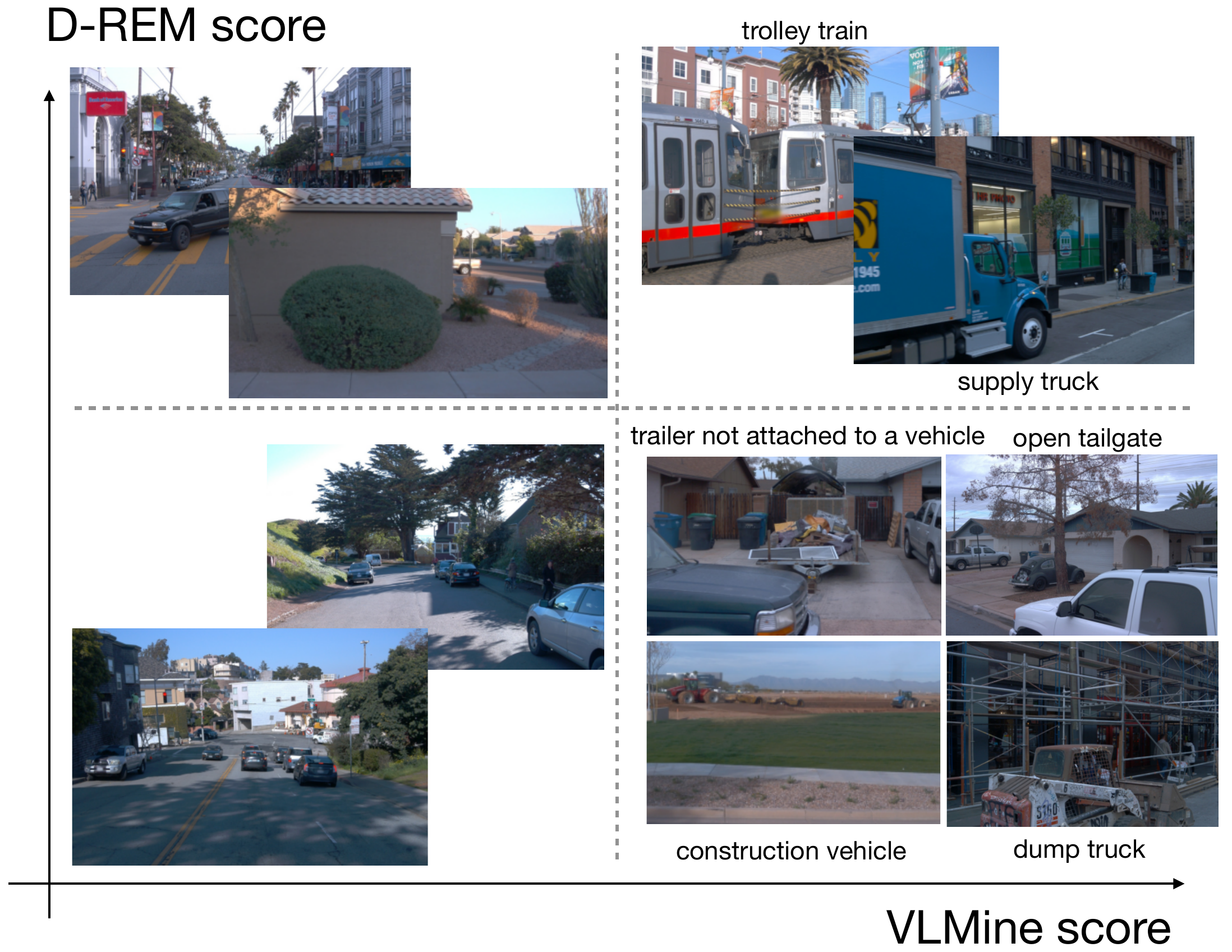}
\caption{Visualization of Pareto and dominated examples from the Waymo Open Dataset. Images are cropped for readability.} \label{fig:pareto_visualization_vehicle}
\end{figure*}

Figure \ref{fig:pareto_visualization_vehicle} illustrates how the signal from VLMine differs from traditional uncertainty-based approaches like D-REM. We depict four types of examples corresponding to long-tail vehicles. Three are Pareto examples where at least one of the mining methods gives the example a high score. The other is a dominated example where both techniques give the example a low score. The examples that have a high novelty score for both D-REM and VLMine are indeed long-tail, such as the trolley train and supply truck. D-REM will often score challenging yet common vehicles highly when VLMine gives them a small novelty score. We obverse that large vehicles tend to be given a higher uncertainty by D-REM. As a result, typical examples with a low D-REM scores but a high VLMine scores are regular-sized but uncommon vehicles, such as a small-sized dump truck or a detached trailer. The examples where both scores are low are common low-complexity scenes with few objects.

\subsection{Ablation Studies} \label{appendix: ablation}

\subsubsection{Prompt Design}
\begin{figure}
  \centering
  \includegraphics[width=0.8\linewidth]{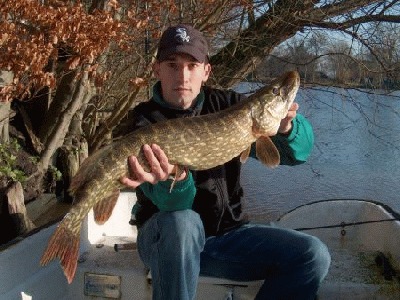}
    \caption{Example of an image from ImageNet-LT.}
    \label{fig:imagenet_example}
\end{figure}

\begin{figure*}[t]
\centering
\includegraphics[width=0.49\linewidth]{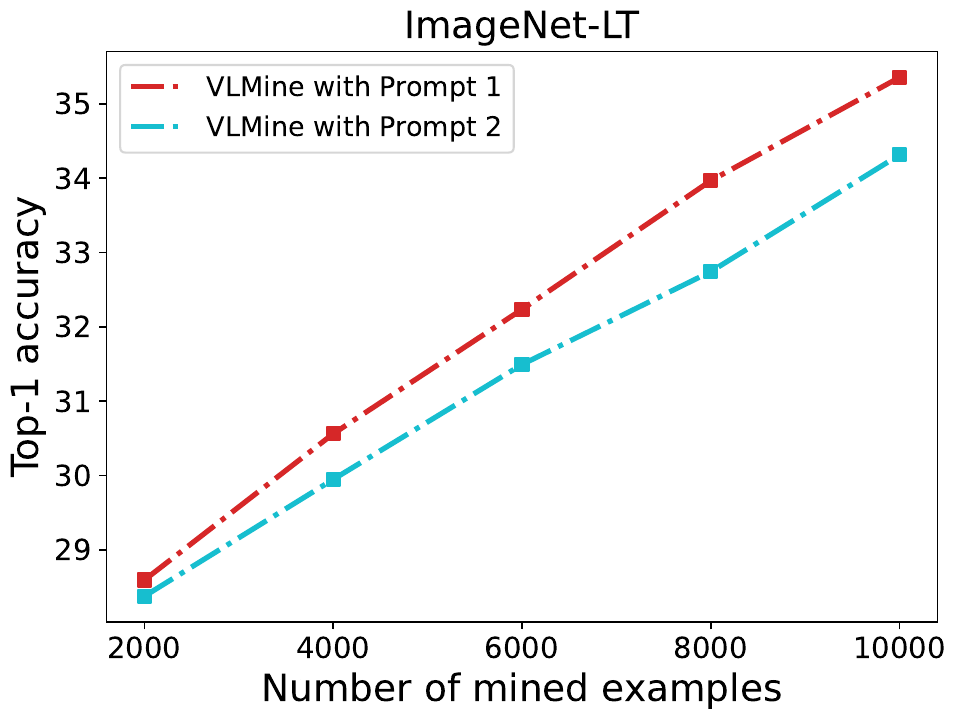}
\includegraphics[width=0.49\linewidth]{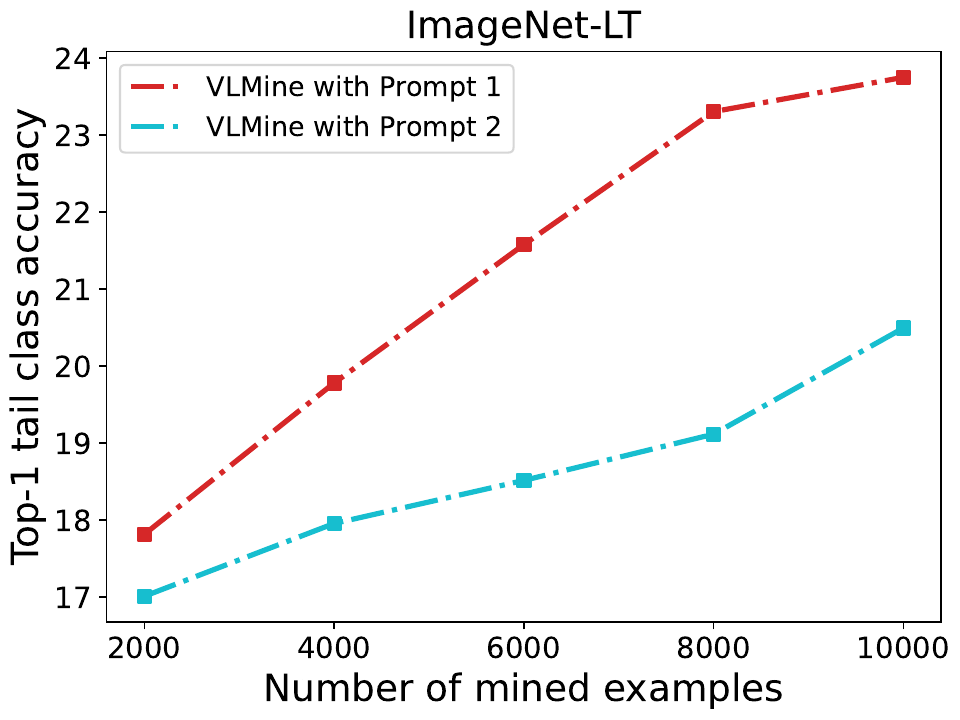}
\caption{Ablation study on prompt engineering with ImageNet-LT.}
\label{fig:ablation_prompt}
\end{figure*}

We study how the design of the prompt influences the final results. For this study, we use ImageNet-LT. For our experiments in Section \ref{sec: 2d_img_experiment}, we prompt the VLM to provide a description of the image with {\ttfamily{What are the possible classes for this image? Give three possible answers}}. We find such a prompt, which we refer to as Prompt 1, will force the model to produce more informative descriptions of the images and thus benefits the data mining. To illustrate this, we conduct an experiment with the following prompt: {\ttfamily{Describe the object in the image}}, which we refer to as Prompt 2. Here are the outputs of the VLM when the two prompts are used for the image shown in Figure \ref{fig:imagenet_example}.
\begin{itemize}
    \item Prompt 1: ``1. Sports and Recreation: The image features a man holding a large fish, which could be a trophy catch from a fishing trip. This suggests that the man is likely an avid fisherman or enjoys spending time outdoors engaging in recreational activities. 2. Nature and Wildlife: The image showcases a man holding a fish, possibly a walleye, in a natural setting. This could be a representation of the beauty of nature and the importance of preserving and protecting aquatic ecosystems. 3. Fishing and Angling: The man is holding a fish, which is a common subject in fishing and angling-related images. This image could be used to promote fishing trips, fishing equipment, or simply to showcase the thrill of catching a large fish.''
    \item Prompt 2: ``The object in the image is a large fish, specifically a muskie, which is being held by a man.''
\end{itemize}
Figure \ref{fig:ablation_prompt} shows the classification accuracy when the different prompts are used. We observe consistent improvement when using Prompt 1, and the improvements are significant for the tail classes.

\begin{figure*}
\begin{subfigure}{.32\textwidth}
  \centering
  \includegraphics[width=\linewidth]{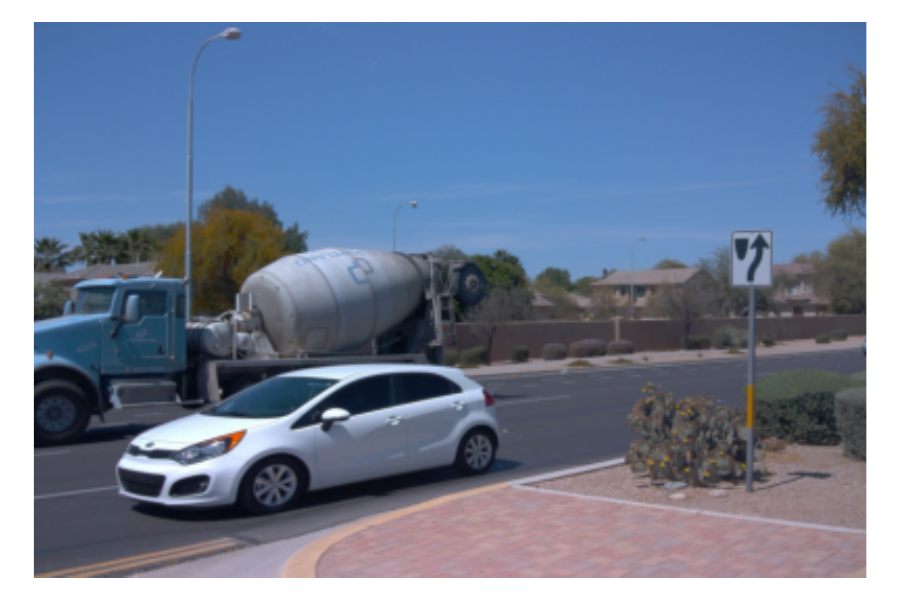}
  \caption{concrete mixer truck}
\end{subfigure}%
\begin{subfigure}{.32\textwidth}
  \centering
  \includegraphics[width=\linewidth]{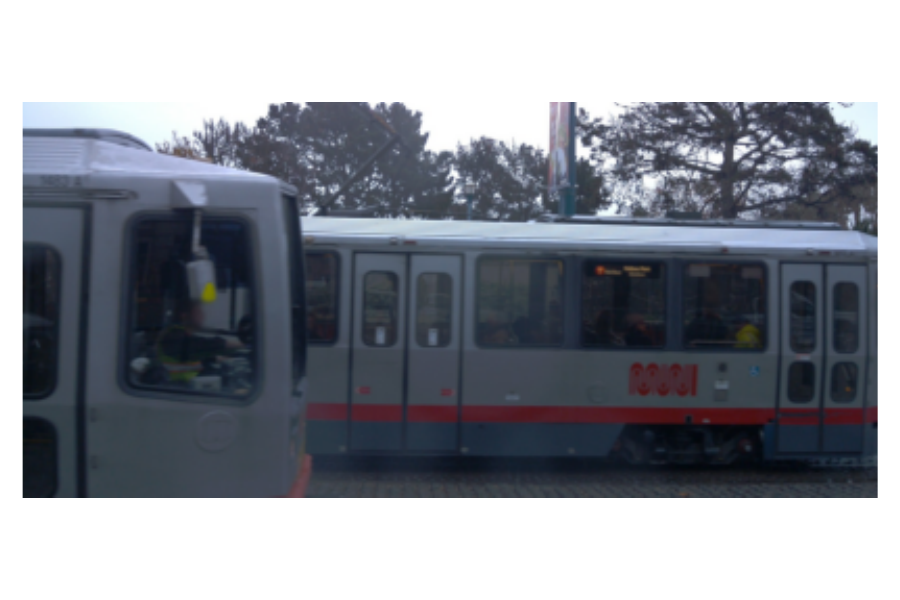}
  \caption{double-length subway train}
\end{subfigure}
\begin{subfigure}{.32\textwidth}
  \centering
  \includegraphics[width=\linewidth]{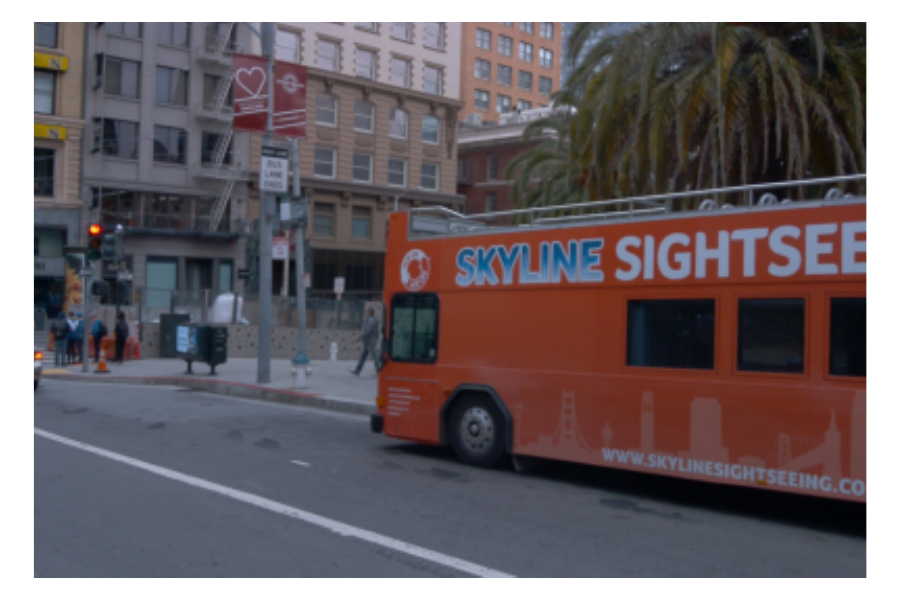}
  \caption{open-top sightseeing bus}
\end{subfigure}
\begin{subfigure}{.32\textwidth}
  \centering
  \includegraphics[width=\linewidth]{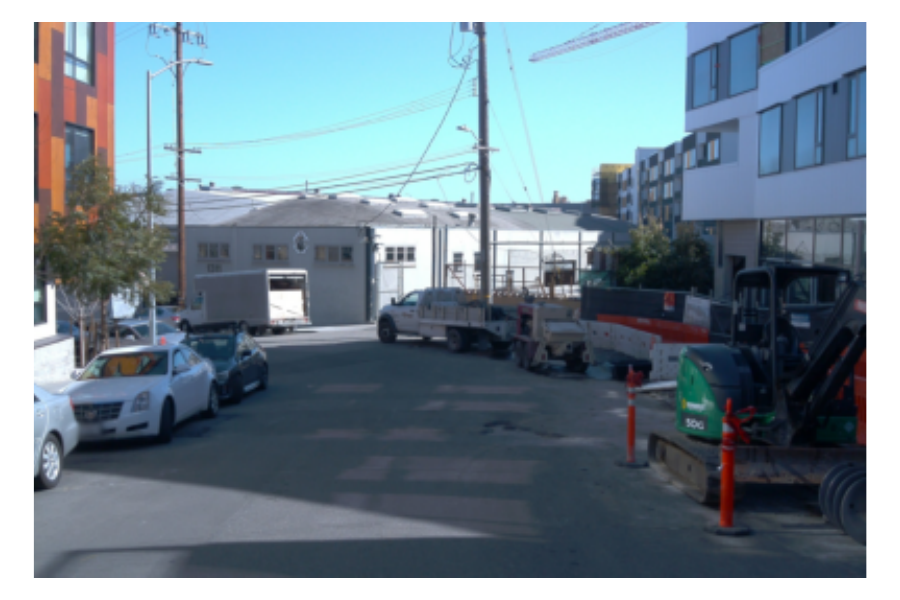}
  \caption{excavator}
\end{subfigure}
\begin{subfigure}{.32\textwidth}
  \centering
  \includegraphics[width=\linewidth]{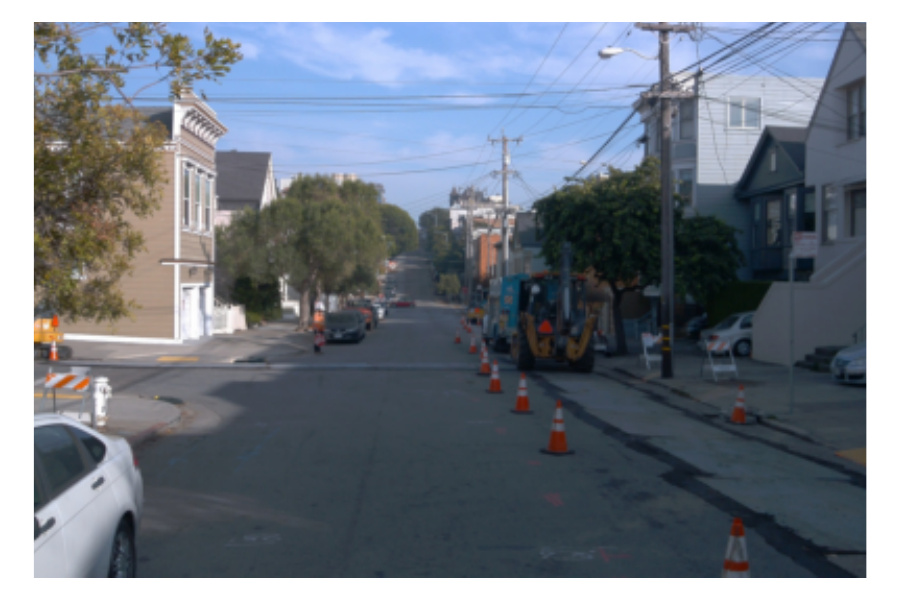}
  \caption{construction vehicle}
\end{subfigure}
\begin{subfigure}{.32\textwidth}
  \centering
  \includegraphics[width=\linewidth]{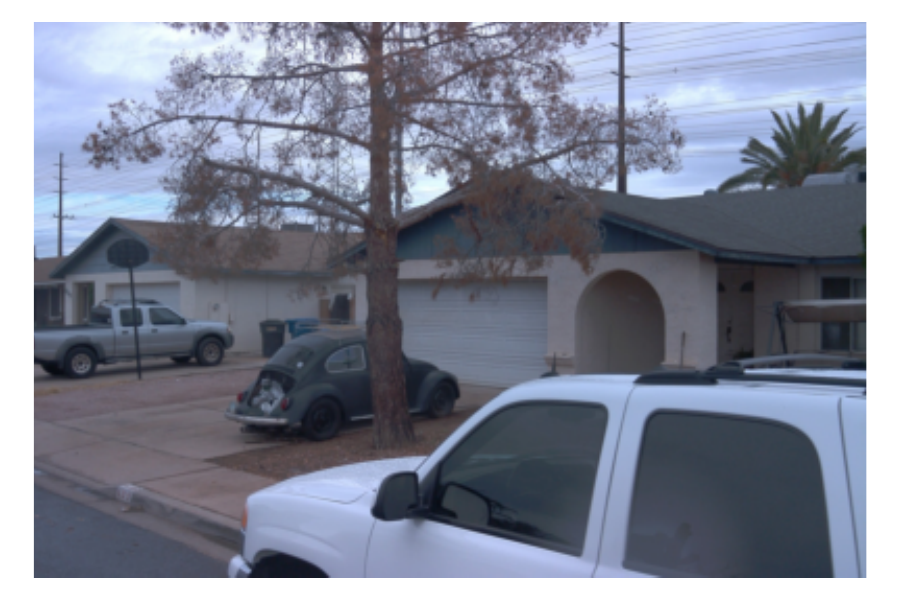}
  \caption{tailgate open}
\end{subfigure}
\begin{subfigure}{.32\textwidth}
  \centering
  \includegraphics[width=\linewidth]{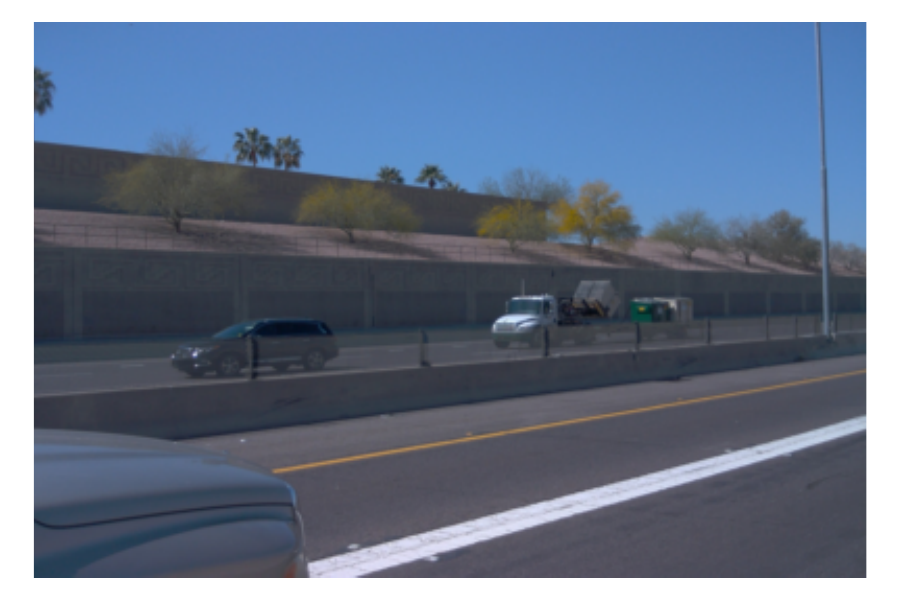}
  \caption{truck with open tailgate}
\end{subfigure}
\begin{subfigure}{.32\textwidth}
  \centering
  \includegraphics[width=\linewidth]{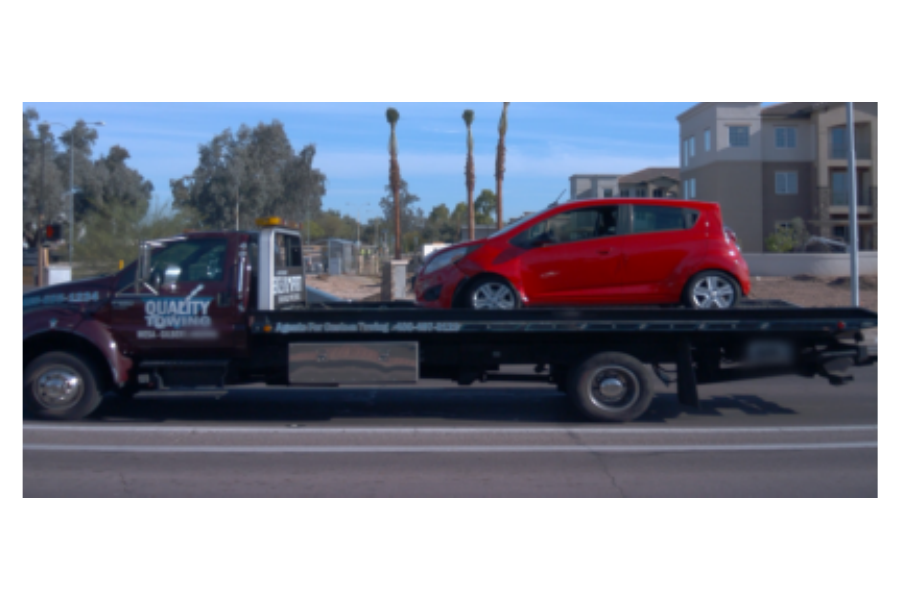}
  \caption{tow truck}
\end{subfigure}
\begin{subfigure}{.32\textwidth}
  \centering
  \includegraphics[width=\linewidth]{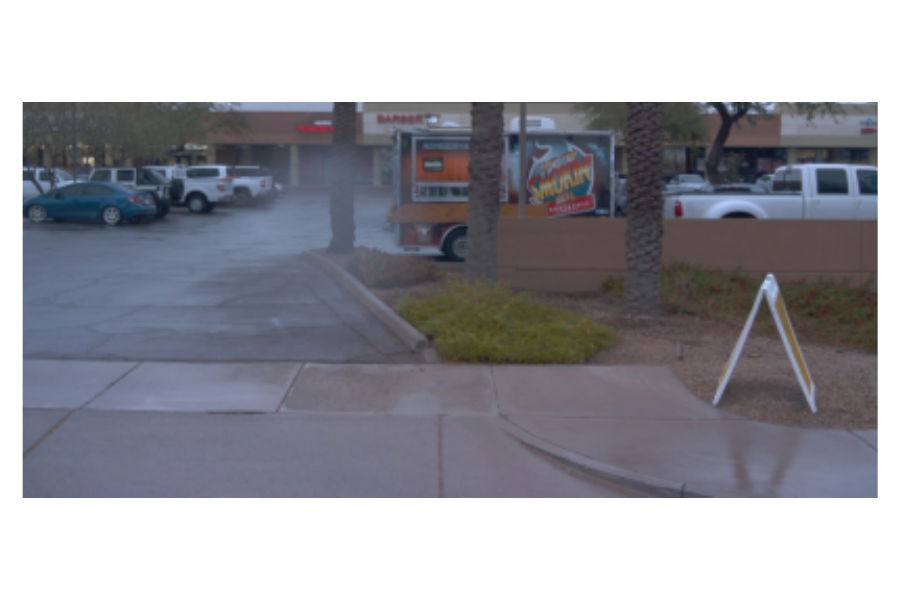}
  \caption{truck carrying advertisement}
\end{subfigure}

\caption{Example images from the Waymo Open Dataset and their keywords.}
\label{fig:vlm_illustration}
\end{figure*}

\subsubsection{Integrating Novelty Scores} \label{appendix: linear combination}

\begin{figure*}[t]
\centering
\includegraphics[width=0.49\linewidth]{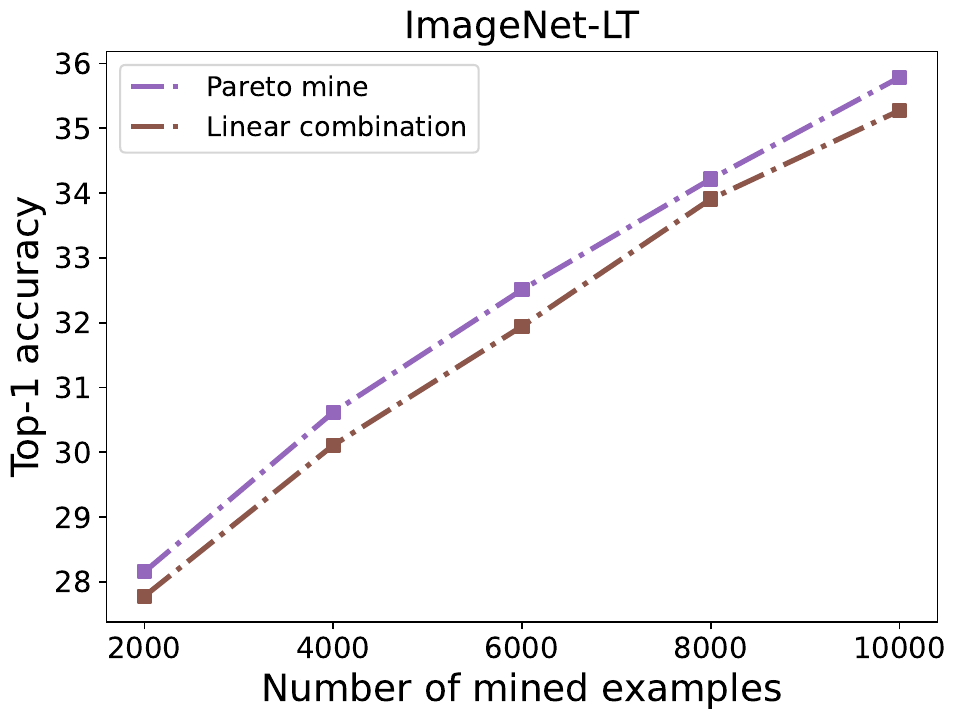}
\includegraphics[width=0.49\linewidth]{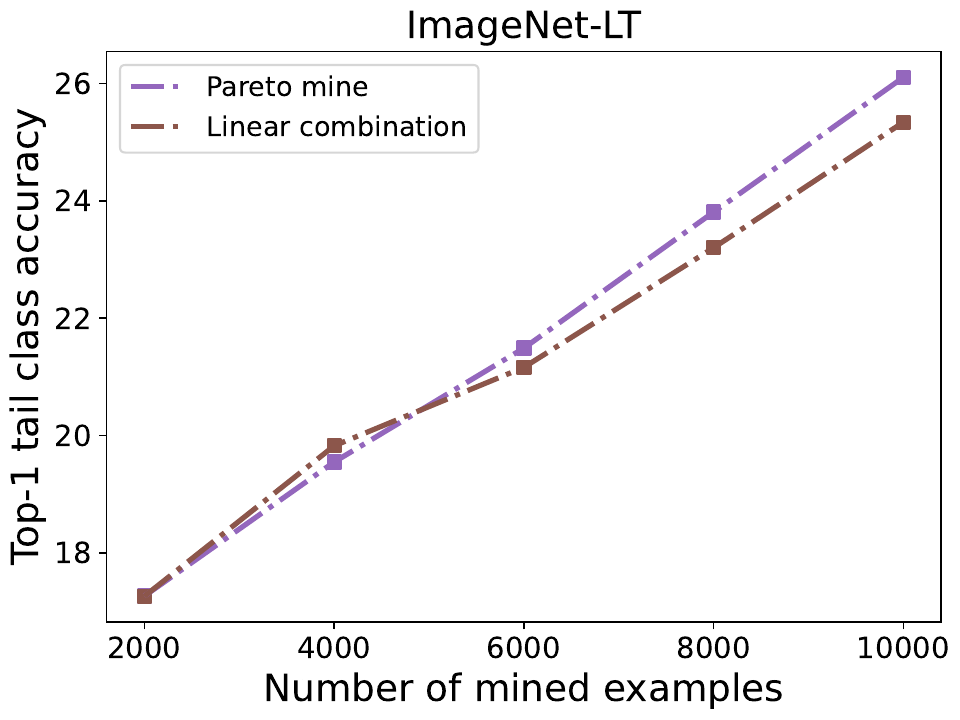}
\caption{Comparing Pareto mining to a linear combination on ImageNet-LT.}
\label{fig:ablation_linear}
\end{figure*}

The proposed Pareto mining provides a simple and principled mechanism to integrate the long-tail signals from multiple algorithms. A linear combination of the novelty scores is another way to leverage multiple signals. In this section, we compare Pareto mining with linear combination. Since the scores from different algorithms might have different statistics, we first normalized all the scores to have a mean of zero and a standard deviation of one. We combine all novelty scores for a particular example by averaging the normalized scores. Figure \ref{fig:ablation_linear} summarizes our results. Overall, we observe that Pareto mining consistently outperforms the linear combination.



\subsection{Remark on One-step Keywords Generation} \label{appendix: one-step}
It is noteworthy that we use a two-step approach to generate keywords for each example. We start by obtaining a description of the example by prompting the VLM, and then in the second step, we extract the keywords from the description using a rule-based approach or a LLM. Alternatively, we can directly ask the VLM to describe the example using keywords. However, we find that the VLM might output non-informative keywords, miss important objects, or generate inconsistent text. For example, using the following prompt: {\ttfamily{List the uncommon or abnormal vehicles, pedestrians and cyclists related to traffic in this image. Return only the keywords in the following format: keyword1, keyword2, etc.}} gives the subsequent results. For Figure  \ref{fig:vlm_illustration}a, the output is repetitive and inconsistent: ``Concrete truck, car, cement mixer, cement truck, white car, concrete truck, cement truck, white car, concrete truck, cement truck, ...'' For Figure  \ref{fig:vlm_illustration}b, important objects are missed: ``None,'' and for Figure  \ref{fig:vlm_illustration}c, keywords are not descriptive: ``Bus.''
It is possible that with a better VLM, we might reconsider one-step keyword generation, but we will leave this for future work.

\begin{figure}
  \centering
  \includegraphics[width=\linewidth]{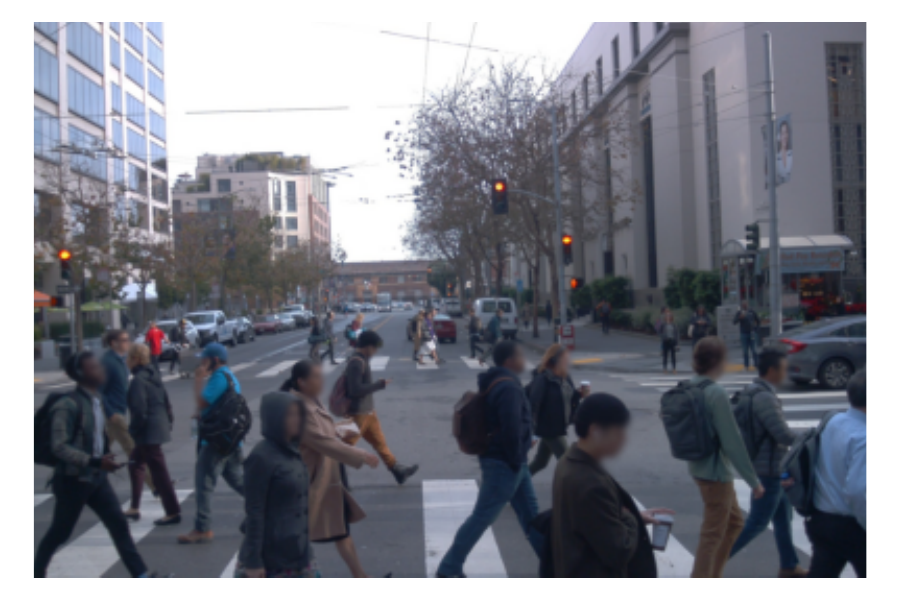}
    \caption{Example of a camera image from the Waymo Open Dataset.}
    \label{fig:wod_example}
\end{figure}

\subsection{Remarks on Hallucination} \label{appendix: hallucination}

The VLM may hallucinate on some input images, but we find that hallucinations often have repeating patterns. For example, a typical case is to hallucinate the presence of a dog when the image shows a busy city street. Below is an example of hallucination when Figure \ref{fig:wod_example} is used as the input image: ``In the image, there are several cars and a bus, which are common vehicles on the road. However, there is also a dog crossing the street along with the group of pedestrians, which is an unusual sight. Additionally, there are multiple bicycles in the scene, including one with a person riding it, which is another uncommon mode of transportation in this particular scene. The presence of these unconventional vehicles and pedestrians creates a unique and lively atmosphere in the busy city street.''

Since such hallucinations share common patterns, they will appear a considerable amount of times when describing the images of a typical autonomous driving dataset. We argue that our approach is robust to these types of hallucinations for two reasons: if the hallucination is not related to the traffic object we are interested in, it will be filtered by the LLM; even if it is related to traffic objects, since it is repeated multiple times, the keywords that come from hallucination will have a relatively high frequency. Of course hallucination is not ideally, but we expect VLMine to improve as the underlining VLM improves.